\documentclass[runningheads,orivec]{llncs}
\usepackage[T1]{fontenc}

\usepackage{graphicx}

\usepackage{amsmath}      
\RequirePackage{amsmath}
\usepackage{booktabs}
\usepackage{xcolor}
\usepackage{amsfonts}      
\usepackage{nicefrac}       

\usepackage{microtype}      
\usepackage{pifont}
\usepackage{adjustbox}
\usepackage{wrapfig}
\usepackage{multirow}
\usepackage{algorithm,algpseudocode}
\usepackage{rotating}
\usepackage{lscape}
\usepackage{longtable}
\usepackage{amssymb}
\usepackage{url}
\usepackage[symbol]{footmisc}

\begin{document}
\title{WSS-CL: Weight Saliency Soft-Guided Contrastive Learning for Efficient Machine Unlearning Image Classification}

\authorrunning{Thang Duc Tran and Thai Hoang Le}

\titlerunning{Weight Saliency Soft-Guided Contrastive Learning for Machine Unlearning}

\author{Thang Duc Tran\inst{1,2}\orcidID{0009-0007-0823-4150}, \and \\ 
Thai Hoang Le\inst{1,2,}\thanks{Corresponding author: lhthai@fit.hcmus.edu.vn} \orcidID{0000-0002-6801-1924}}

\institute{Faculty of Information Technology, University of Science, Ho Chi Minh City, Vietnam. \and
Vietnam National University, Ho Chi Minh City, Vietnam. \\
% \email{23C11047@student.hcmus.edu.vn}\\
}

\maketitle              % typeset the header of the contribution
\begin{abstract}
Machine unlearning, the efficient deletion of the impact of specific data in a trained model, remains a challenging problem. Current machine unlearning approaches that focus primarily on data-centric or weight-based strategies frequently encounter challenges in achieving precise unlearning, maintaining stability, and ensuring applicability across diverse domains. In this work, we introduce a new two-phase efficient machine unlearning method for image classification, in terms of weight saliency, leveraging weight saliency to focus the unlearning process on critical model parameters. Our method is called \textit{weight saliency soft-guided contrastive learning for efficient machine unlearning image classification (WSS-CL)}, which significantly narrows the performance gap with "exact" unlearning. First, the forgetting stage maximizes Kullback–Leibler divergence between output logits and aggregated pseudo-labels for efficient forgetting in logit space. Next, the adversarial fine-tuning stage introduces contrastive learning in a self-supervised manner. By using scaled feature representations, it maximizes the distance between the forgotten and retained data samples in the feature space, with the forgotten and the paired augmented samples acting as positive pairs, while the retained samples act as negative pairs in the contrastive loss computation. Experimental evaluations reveal that our proposed method yields much-improved unlearning efficacy with negligible performance loss compared to state-of-the-art approaches, indicative of its usability in supervised and self-supervised settings.

\end{abstract}

\section{Introduction}

The increasing use of machine learning models in numerous domains, alongside "the right to forget" and data privacy concerns, has placed in the spotlight \cite{bourtoule2021machine} the problem of machine unlearning. Machine unlearning aims to delete effectively and efficiently the contribution of individual training samples in a trained model as if such samples were not experienced at training time. This property is critical for GDPR compliance \cite{voigt2017eu}, delete request processing, and the protection of model prediction accuracy over biases and inaccuracies in data.

Prior work in machine unlearning has involved full retraining, excluding forgetting data, and is computationally infeasible and costly for large datasets and complex models \cite{cao2015towards}. Recently, there have been works with a direction towards efficient technique development, such as approximation unlearns and sharded unlearns \cite{bourtoule2021machine}. There is, nevertheless, a challenge in achieving effective unlearning with zero performance loss and computational cost.

We introduce in this work a new two-step training mechanism for efficient unlearning in image classification, with a direction towards weight saliency and contrastive learning. Our algorithm identifies and updates specific model weights that are most salient to the target forgetting data, rather than retraining in full. Targeted unlearns deliver a significant improvement in efficiency and performance over traditional unlearns.

Our proposed scheme consists of a forgetting stage and an adversarial fine-tuning stage. In the forgetting stage, unlearning in logit space is emulated through Kullback–Leibler divergence maximization between model output logits and aggregated pseudo-labels for forgetting samples. This effectively initiates the "forgetting" mechanism.

The subsequent stage of adversarial fine-tuning then fine-tunes the model with a new forgetting rotation mechanism derived from self-supervised contrastive learning. This mechanism aims to maximize the feature-space distance between representations for samples to forget and samples to remember. Specifically, samples to forget and their augmented variants act as positive pairs, while samples to remember act as negative pairs in a contrastive loss calculation. This effectively "pushes away" the impact of forgotten data. In addition, weight saliency is soft incorporated in the contrastive loss calculation, such that unlearning activity is concentrated in the most relevant model parameters. Experimental evaluations confirm that our scheme attains much-improved unlearning performance with minor performance drop-off in comparison to state-of-the-art alternatives, with its adaptability in both supervised and self-supervised scenarios.

\section{Related Work}

\subsection{Machine Unlearning in Image Classification}

Machine unlearning (MU) seeks to remove specific data points or even an entire class's contribution from trained models, overcoming privacy-related concerns such as membership inference attacks \cite{Ginart2019}, \cite{Neel2021}. The most common practice, \textbf{exact unlearning}, relearns the model in its entirety starting with a reduced dataset \( D_r \). SISA \cite{Bourtoule2021} and ARCANE \cite{Yan2022}-like methodologies partition data to save computational costs. However, such exact techniques involve high computational expense and sensitivity to partitioning techniques.

To overcome such constraints, \textbf{approximate unlearning} techniques have been proposed \cite{golatkar2020eternal}. Techniques such as fine-tuning (FT) \cite{golatkar2020eternal}, gradient ascent (GA) \cite{graves2021amnesiac}, fisher forgetting (FF) \cite{golatkar2020eternal}, and influence unlearning (IU) \cite{izzo2021approximate} strike a balance between efficiency and effectiveness. Despite such efficiency, these techniques fall short in terms of unlearning effectiveness, performance in a variety of scenarios, and restricted usability in contexts other than supervised learning.

\textbf{Exact unlearning techniques} ensure full erasure of the contribution of the forgetting dataset \( D_f \) from a trained model. The SISA \cite{Bourtoule2021}-like partition scheme shreds data and relearns only the impacted shreds in the case of an unlearning request, reducing computational expenses compared to full retraining. Likewise, ARCANE \cite{Yan2022} employs partitioning at the class level to support full-class unlearning through relearning of applicable parts. Despite such success, these techniques involve high computational expenses and require efficient partitioning.

\textbf{Approximate unlearning techniques} present a pragmatic alternative by minimizing the impact of \( D_f \) without full retraining. Fine-tuning (FT) \cite{golatkar2020eternal}, compared to retraining, involves fine-tuning the initial model \( \theta \) over the retention dataset \( D_r \) for a short training period, generating the updated model \( \theta_u \). This process induces catastrophic forgetting of the forgetting dataset \( D_f \), a behavior that is widespread in continual training environments. Gradient ascent (GA) \cite{graves2021amnesiac} counteracts training over \( D_f \) by adding respective gradients to the initial model parameters \( \theta_o \). This effectively raises the loss for such scheduled erasure examples, thereby unlearning them.

Fisher forgetting (FF) \cite{golatkar2020eternal} employs an additive Gaussian perturbative scheme to perturb \( \theta_o \) towards exact unlearning. The Gaussian perturbative distribution is defined with zero mean and covariance with the fourth root of the Fisher information matrix concerning \( \theta_o \) over \( D_r \). Nonetheless, estimating the Fisher information matrix incurs high computational costs and less parallel efficiency concerning GPU architectures compared to alternative unlearning algorithms \cite{golatkar2020eternal}. Influence unlearning (IU) \cite{izzo2021approximate} utilizes the theory of influence functions \cite{cook1982residuals} to estimate the impact of erasing specific training examples on model parameters. It estimates the transition between \( \theta_o \) and \( \theta_u \) as \( \theta_u - \theta_o \). IU is related to \( \epsilon \)-\( \delta \) forgetting \cite{wang2022federated}, but it tends to require additional model and training scheme assumptions \cite{wang2022federated}. While these methodologies maximize computational efficiency, they fall short in unlearning effectiveness and robustness across various scenarios.

\subsection{Weight Saliency in Machine Unlearning}

Saliency analysis deepens model interpretability in machine learning by discovering significant input features or model parameters. Saliency maps of inputs, including pixel-space sensitivity maps \cite{Simonyan2013} and class-discriminative localization techniques \cite{Zhou2016}, have become prevalent in model explanation. In addition, attribution techniques for data \cite{Park2023} estimate the individual contribution of samples of data towards model prediction, contributing to model debugging \cite{Ilyas2022}, efficient training \cite{Xie2023}, and generalization improvement \cite{Jain2023}.

Whereas weight saliency is less mature, techniques in weight sparsity \cite{Han2015}, for instance, in weight pruning, detect and preserve the most significant weights, boosting model efficiency and generalization. In NLP, model editing techniques \cite{Dai2021} target individual model weights for updating and deleting specific knowledge, similar to weight saliency in MU.

The most recent work in MU utilizes weight saliency for unlearning effectiveness and efficiency improvement. By discovering and updating the most significant model weights, it can delete specific data with high accuracy and fewer retrainings. Techniques such as weight pruning \cite{Han2015} induce sparsity, preserving critical weights while maintaining model performance. In NLP, model editing techniques \cite{Patil2023} introduce model weight pruning for specific and effective deletion of knowledge, in consonance with MU requirements. One of the most advanced saliency mask techniques is salUn \cite{fan2023salun}. SalUn is a first-principles MU approach that can effectively remove the effects of forgetting data, classes, or concepts in both classification and image generation tasks.

\subsection{Our Contribution}

Building on existing work, our contribution proposes soft saliency of weights in a two-stage training scheme for MU. The scheme involves a forgetting stage and an adversarial fine-tuning stage, unlearning specific effective weights in that direction. In addition, we use self-supervised contrastive learning in the forgetting rotation, such that forgetting samples become uniformly scattered in feature space and decoupled from preserved samples. With a dual objective, efficiency and effectiveness in unlearning for both deletion of samples and classes are maximized, outpacing current empirical and certified alternatives in model utility maintenance and computational efficiency.

Our contribution addresses the necessity for unlearning in machines for data privacy by erasing specific data impacts in trained models. Complete erasure with exact approaches involves computational infeasibility in practice. Approximation and certification techniques for unlearning have increased usability but lack effectiveness, stability, and extensibility. With saliency of weights and through integration with complex training, our contribution advances in providing an effective, efficient, and powerful unlearning mechanism for a variety of unlearning scenarios.

\section{Proposed Method}

\begin{figure*}[t]
    \centering
    \includegraphics[width=1\textwidth]{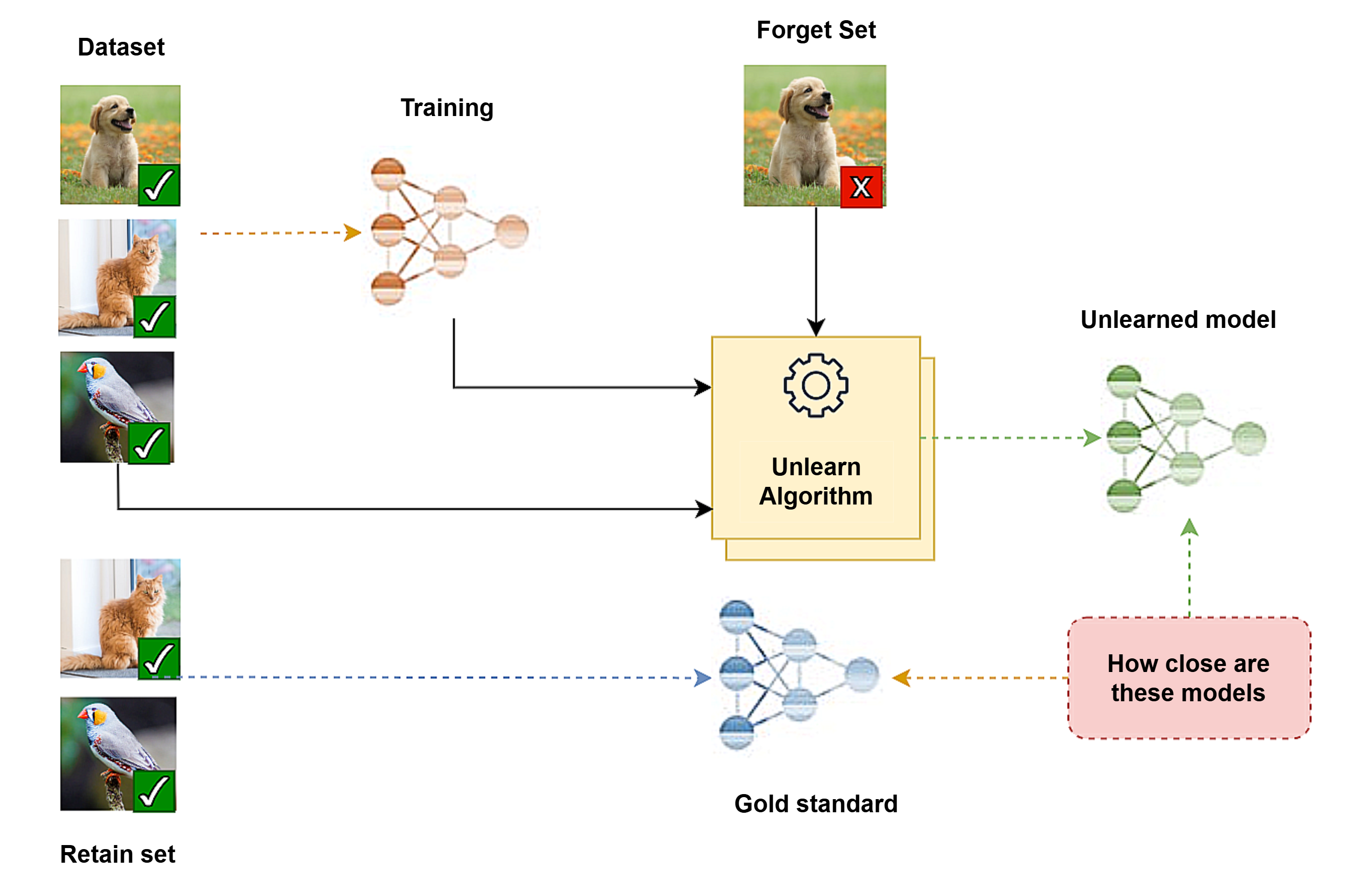}
    \caption{A conceptual illustration of the relearning process. The image provides a conceptual visualization of how data from the "forgetting set" is removed from the pre-trained model, creating an "unlearned" model, and how it compares to a "gold standard" model trained without the forgetting set.}
    \label{MU_arch}
\end{figure*}

\subsection{Problem Statement}

Formally, the MU problem can be posed in terms of figure \ref{MU_arch} below:
\begin{itemize}
    \item \textbf{Given:}
    \begin{itemize}
        \item Pre-trained model $ \theta_o $ trained over dataset $D = \{z_i\}_{i=1}^{N}$.
        \item Forgetting dataset $D_f \subseteq D$.
        \item Rest dataset $D_r = D \setminus D_f$.
    \end{itemize}
    \item \textbf{Find:}
    \begin{itemize}
        \item An unlearned model $ \theta_u $ derived from $ \theta_o $ such that its contribution of $D_f$ is removed.
    \end{itemize}
    \item \textbf{Such That:}
    \begin{itemize}
        \item $ \theta_u $ closely emulates performance of $ \theta_{\text{Retrain}} $, retrained from scratch over $D_r$.
        \item Computationally efficient unlearning in comparison with retraining with high accuracy.
        \item High fidelity unlearned model over $D_r$ and generalizes over new samples.
    \end{itemize}
\end{itemize}
Our proposed solution to such a problem entails a two-step training scheme with accompaniment with weight saliency for improvement in unlearning accuracy and efficiency at a loss in model performance.

\subsection{Method Overview}

\begin{figure*}[t]
    \centering
    \includegraphics[width=\textwidth]{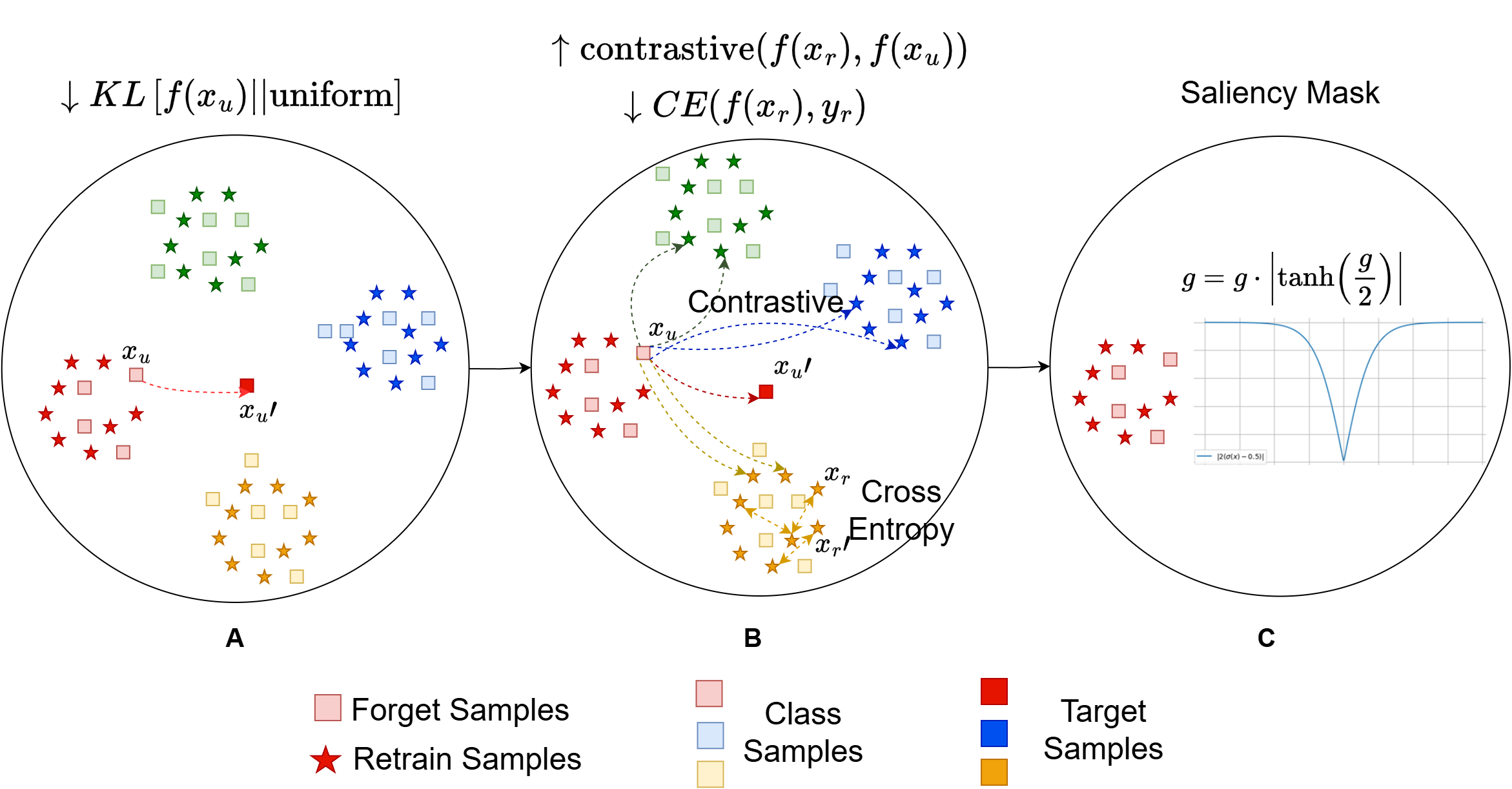}
    \caption{Overview of the proposed machine unlearning architecture. The two-stage training process consists of the forgetting phase and the adversarial fine-tuning phase, integrated with saliency masking and contrastive learning techniques.}
    \label{architecture}
\end{figure*}

In this section, we present our proposed scheme for machine unlearning (MU) as shown in the Fig. \ref{architecture}, which can effectively remove the contribution of individual samples or even entire classes in a trained model with no loss in its overall performance. Our scheme employs a two-step training mechanism that integrates Kullback-Leibler (KL) divergence optimization in Fig. \ref{architecture} A , contrastive learning in Fig. \ref{architecture} B, and saliency masking in Fig. \ref{architecture} C, resulting in improvements in both effectiveness and computational efficiency in unlearning.

The first stage consists of a forgetting phase, in which model training is conducted only over the forgetting dataset $D_f$. In this stage, KL divergence between model output logits and a uniform pseudo-class distribution is optimized. Optimizing KL divergence encourages the model to produce uniform class probabilities over the forgetting data, effectively deleting its contribution and ceasing to be predictive of such data samples.

The second stage consists of an adversarial fine-tuning stage, in which model refinement is conducted through contrastive learning over the forgetting dataset $D_f$ and cross-entropy loss over the retention dataset $D_r$. In this stage, a saliency mask is leveraged for a selective update of model weights that are most critical for forgetting specific data. By employing cross-entropy loss over $D_r$, model performance over $D_r$ is assured, and contrastive learning aids in effective unlearning over $D_r$. This two-way focused scheme not only aids in unlearning effectiveness but also in maintaining model performance over $D_r$.

By directing unlearning at specific, salient model weights rather than at the model in general, our work realizes significant efficiency and performance gains. In the following, individual components of our work, including the KL-divergence-based forgetting stage, saliency-mask integration, and contrastive learning, will be discussed in detail.

\subsection{Kullback-Leibler (KL) Divergence}

\label{sec:kl_divergence}

Our primary objective is to adapt the model in such a manner that, when it is trained with examples in the forgetting set \( D_f \), it generates a uniform distribution over all classes. By doing this, we cancel out any predictive bias in the model for any specific classes related to the forgotten examples. KL divergence is a distance measure between one distribution and another, a fundamental concept in information theory \cite{kullback1951information}. In our case, we utilize KL divergence in such a way that the unlearning of \( D_f \) can be facilitated by minimizing the distance between the model output and a uniform distribution in logit space.

Formally, during the forgetting process, we minimize the target using the following equation 1.

\begin{align}
\mathcal{L}_{KL} &= \text{KL}\left( P_{\theta_o}(y \mid x) \,\|\, P_{\text{uniform}}(y) \right),
\end{align}

Where $P_{\theta_o}(y \mid x)$ is the model output predicted probability distribution over classes for an input $x$ in the forgetting set $D_f$. $P_{\text{uniform}}(y)$ is a uniform pseudo-label distribution, such that for any class $y$, $P_{\text{uniform}}(y) = \frac{1}{K}$, with $K$ representing the number of classes. By minimizing $ \mathcal{L}_{KL} $, the model output for $D_f$ ultimately has a uniform distribution, effectively erasing any mapping between $D_f$ and individual classes. Thus, the model output for $D_f$ ultimately has a uniform uncertainty over all classes, and unlearning is achieved.

\subsection{Saliency Masking}

To enhance efficiency and effectiveness in unlearning, the most important model parameters, a saliency masking mechanism similar to \cite{fan2023salun}, are utilized. As an improvement over the saliency map through gradients \cite{adebayo2018sanity}, our mechanism identifies salient weights relevant for unlearning. Let model parameters be denoted as $ \theta $. Algorithm steps for saliency identification and selective update of model parameters for unlearning follow below:

\textbf{Saliency Identification}. Deduce forgetting loss $ L_{\text{forget}} $'s gradient with regard to model parameters $ \theta $, based on equation 2 as follows.
\begin{align}
g_\theta = \frac{\partial L_{\text{forget}}}{\partial \theta},
\end{align}
Equation 3 represents a simple and traditional one for estimating importance for a parameter is through its absolute value of its respective gradient.
\begin{align}
s = |g_\theta|,
\end{align}
For a soft estimation, a soft saliency value through sigmoid function calculation is considered, like the following equation 4.
\begin{align}
s = \left|2\left(\sigma(g_\theta) - 0.5\right)\right|,
\end{align}
Here, $ \sigma(\cdot) $ is for the sigmoid function, mapping gradients onto a continuous range with added weightage for values with high deviation from zero, providing a soft estimation even for minor discrepancies in importance between parameters.

Alternatively, by leveraging the soft saliency values, we can set it directly based on equation 5 as follows:
\begin{align}
M_{\theta} = s_{\theta},
\end{align}
Hence, providing a continuous modality for regulating updates in gradients. For update phases in $ \theta $, computed gradients are updated through element-wise multiplication with a saliency mask $ M $, expressed in equation 6 as follows:
\begin{align}
g = M \odot g,
\end{align}
Here, $ \odot $ is for element-wise multiplication. The update, in such a manner, will target focused updates in salience-masked $ \theta $'s, effectively channeling unlearning towards important model weights.

\subsection{Contrastive Learning for Forgetting Rotation}

To improve effectiveness in unlearning, a specific mechanism of contrastive learning \cite{khosla2020supervised} is incorporated into our stage of adversarial fine-tuning in our machine unlearning system. For forgetting samples, a specific mechanism, forgetting rotation, is employed for partitioning the representations of forgetting and retention datasets in feature space, such that the model learns to forget specific information while maintaining strong performance for kept-back samples. Two standalone modules constitute our solution: Forgetting sample contrast and balanced contrast with cross-entropy loss.

The forgetting sample contrast module is used to minimize model overreliance on the forgetting dataset \( D_f \). For any forgetting sample \( x \in D_f \), an augmented counterpart $x'$ is generated, forming a positive pair \((x, x')\) for contrastive loss computation. The pairs have similar representations, and the model clusters feature representations closely together in latent space. In parallel, all samples in the retention batch $ D_r $ serve as negative samples. The contrastive loss function aims to minimize the distance between $x$ and $x'$, such that augmented representations of the same forgetting sample form closely grouped clusters, and maximize the distance between $x$ (and $x'$) and any kept-back sample in $D_r$, effectively moving forgetting samples apart from kept-back samples.

The contrastive loss for forgetting samples mathematically can be represented by equation 7 as follows.
\begin{align}
\mathcal{L}_{\text{forget}} = -\log \frac{\exp\left(\frac{\text{sim}(x, x')}{\tau}\right)}{\exp\left(\frac{\text{sim}(x, x')}{\tau}\right) + \sum_{x_r \in D_r} \exp\left(\frac{\text{sim}(x, x_r)}{\tau}\right)},
\end{align}
Where $ \text{sim}(\cdot, \cdot) $ is a function for a specific form of similarity, such as cosine similarity. \( \tau \) is a temperature hyperparameter for logits' scaling, and in our solution, it is taken with value $ \tau = 1.4 $.

This loss function encourages feature representations learned by the model to become invariant to forgetting samples' augmentations and differentiate them from retained samples, thus enabling effective unlearning.

Whereas forgetting sample contrast aims to distinguish forgetting samples from retained samples, balanced contrast ensures that the performance and integrity of the retention dataset \( D_r \) are not compromised. To make this a reality, we use a cross-entropy loss function specifically designed for the retention dataset. This deliberate switch between contrastive loss and cross-entropy loss plays a dual role. The cross-entropy loss encourages the model to maintain correct predictions for retained samples, thereby preserving its performance in \( D_r \). By utilizing cross-entropy loss for \( D_r \), we prevent the model from driving retained samples apart, a move that would reduce classification performance.

The cross-entropy loss by equation 8 is as follows.
\begin{align}
\mathcal{L}_{\text{CE}} = -\sum_{k=1}^{K} y_{r,k} \log \left( \frac{\exp(x_{r,k}) }{\sum_{j=1}^{K} \exp(x_{r,j})} \right ),
\end{align}
Where \( K \) is the number of classes, and \( y_{r,k} \) is the binary indicator (0 or 1) if class label \( k \) is the correct classification of \( x_r \).

Our new MU framework employs a synergy of KL divergence optimization, saliency masking, and contrastive learning in a two-stage training framework to efficiently and effectively unlearn. Our approach, by selectively targeting certain model weights and employing forgetting rotation, ensures the elimination of unwanted data influences without affecting the performance and generalization ability of the model.

\section{Experiments}

\subsection{Experiment setups}

\noindent 

In this section, experimental settings for image classification experiments, including datasets, model architectures, unlearning settings, and evaluation metrics for comparing the performance of our proposed unlearning in machines scheme, are addressed.

\textbf{Models and datasets}. In image classification experiments, the CIFAR-10 \cite{krizhevsky2009learning} dataset is typically used with the model architecture of ResNet-18 \cite{he2016deep}. With this configuration, one can explore randomness in forgetting data and assess unlearning effectiveness over a renowned benchmark. To extend the evaluation, we also use an additional dataset, namely the CIFAR-100 \cite{krizhevsky2009learning}, with the model architecture of ResNet-18.

\textbf{Unlearning sets and methods}.In our experiments, we address both forgetting scenarios for a target class and randomness in forgetting in image classification. For randomness in forgetting, we remove about 10\% or 50\% of training samples, and selection is performed several times for statistical purposes. In the case of forgetting for a target class, we remove samples belonging to that class and unlearn them in a trained model.

\textbf{Evaluation metrics}. For performance evaluation of unlearning, we utilize a mix of evaluation metrics adopted from current unlearning work. Unlearning accuracy (UA) and membership inference attack (MIA) are used to test the effect of forgetting the training dataset $D_f$. We use MIA here which is black box membership threshold inference attack. Remaining accuracy (RA) and testing accuracy (TA) are used to evaluate the fidelity and generalization ability of the unlearned model concerning the retention dataset $D_r$. Run-time efficiency (RTE) is used to examine the computational cost of unlearning operations.

\subsection{Experiment results}

In this section, we present our proposed machine unlearning techniques. Since the Kullback-Leibler method is used for all our architectures, we omit its name in the abbreviation. For the Kullback-Leibler architecture combined with contrastive learning, we call it \textit{contrastive learning (CL)}. For the weight saliency architecture combined with contrastive learning, we call it \textit{weight saliency-contrastive learning (WS-CL)}, and finally, for the weight saliency soft architecture combined with contrastive learning, we call it \textit{weight saliency soft-contrastive learning (WSS-CL)}. The methods are compared with eight basic MU techniques: FT, RL, GA, IU, $\ell_1$-Sparse, BS, BE, and SalUn. All experiments have been conducted using the CIFAR-10 dataset and the ResNet-18 model. Additionally, we used the CIFAR-100 dataset with the ResNet-18 model. Two unlearning scenarios, \textbf{10\%} and \textbf{50\%} of uniformly randomly forgotten data, have been taken into consideration in an attempt to assess each technique's robustness and scalability. Performance is measured with six evaluation metrics: Unlearning accuracy (UA), remaining accuracy (RA), test accuracy (TA), membership inference attack (MIA), average gap (Avg. Gap), and runtime efficiency (RTE). The detailed results are presented in Table~\ref{random_forgetting_10_Cifar10}, Table~\ref{random_forgetting_50_Cifar10}, Table~\ref{random_forgetting_10_Cifar100}, and Table~\ref{random_forgetting_50_Cifar100}. The comparison results for specific class forgetting are presented in Table~\ref{two_model_five_methods}.

\begin{table}[htbp]
\caption{
Performance metrics of different MU methods in image classification with 10\% randomly forgotten data on CIFAR-10 using ResNet-18. The absolute performance gap compared to Retrain is given in (•), blue text shows the absolute gap.
}
\label{random_forgetting_10_Cifar10}
\centering
\resizebox{0.8\textwidth}{!}{
\begin{tabular}{c|cccccc}
\toprule[1pt]
\textbf{Method} & \textbf{$UA\uparrow$} & \textbf{$RA\uparrow$} & \textbf{$TA\uparrow$} & \textbf{$MIA\uparrow\downarrow$} & \textbf{$Avg.\;Gap\downarrow$} & \textbf{$RTE\downarrow$} \\
\midrule
\textbf{Retrain}                         & 5.4    & 100.00                               & 94.08                                & 12.88                               & 0.00                                & 43:17   \\
\midrule
\textbf{FT \cite{warnecke2021machineunlearning}}     & 0.63 \textcolor{blue}{(4.77)}   & \textbf{\underline{99.88}} \textcolor{blue}{(0.12)}   & \textbf{\underline{94.06}} \textcolor{blue}{(0.02)}   & 2.70 \textcolor{blue}{(10.18)}     & 3.78                               & 2:22    \\
\textbf{RL \cite{golatkar2020eternal}}              & 7.61 \textcolor{blue}{(2.21)}   & 99.67 \textcolor{blue}{(0.33)}                      & 92.83 \textcolor{blue}{(1.25)}                      & 37.36 \textcolor{blue}{(24.47)}    & 7.07                               & 2:38    \\
\textbf{GA \cite{thudi2022auditableunlearning}}      & 0.69 \textcolor{blue}{(4.71)}   & 99.50 \textcolor{blue}{(0.50)}                      & 94.01 \textcolor{blue}{(0.07)}                      & 1.70 \textcolor{blue}{(11.8)}      & 4.12                               & 0:08    \\
\textbf{IU \cite{koh2017influencefunctions}}         & 1.07 \textcolor{blue}{(4.33)}   & 99.20 \textcolor{blue}{(0.80)}                      & 93.20 \textcolor{blue}{(0.88)}                      & 2.67 \textcolor{blue}{(10.21)}     & 4.06                               & 3:13    \\
\textbf{BE \cite{chen2023boundaryunlearning}}        & 0.59 \textcolor{blue}{(4.81)}   & 99.42 \textcolor{blue}{(0.58)}                      & 93.85 \textcolor{blue}{(0.23)}                      & 7.47 \textcolor{blue}{(5.41)}      & 2.76                               & 0:16    \\
\textbf{BS \cite{chen2023boundaryunlearning}}        & 1.78 \textcolor{blue}{(3.62)}   & 98.29 \textcolor{blue}{(1.71)}                      & 92.69 \textcolor{blue}{(1.39)}                      & 8.96 \textcolor{blue}{(3.92)}      & 2.67                               & 0:26    \\
\textbf{$\ell_1$-Sparse \cite{koh2017influencefunctions}} & 4.19 \textcolor{blue}{(1.21)}   & 97.74 \textcolor{blue}{(2.26)}                      & 91.59 \textcolor{blue}{(2.49)}                      & \textbf{\underline{9.84}} \textcolor{blue}{(3.04)} & 2.26                               & 2:22    \\
\textbf{SalUn \cite{fan2023salun}}                   & 4.18 \textcolor{blue}{(1.22)}   & 99.13 \textcolor{blue}{(0.87)}                      & 93.17 \textcolor{blue}{(0.91)}                      & 18.69 \textcolor{blue}{(5.81)}     & 2.20                               & 2:40    \\
\midrule
\textbf{CL}                              & 4.09 \textcolor{blue}{(1.31)}   & 99.50 \textcolor{blue}{(0.5)}                       & 93.29 \textcolor{blue}{(0.79)}                      & 16.71 \textcolor{blue}{(3.83)}     & 1.60                               & 2:53    \\
\textbf{WS-CL}                           & 4.18 \textcolor{blue}{(1.22)}   & 99.12 \textcolor{blue}{(0.88)}                      & 93.24 \textcolor{blue}{(0.84)}                      & 16.15 \textcolor{blue}{(3.27)}     & 1.55                               & 2:55    \\
\textbf{WSS-CL}                          & \textbf{\underline{4.51}} \textcolor{blue}{(0.89)} & 99.39 \textcolor{blue}{(0.61)}                      & 93.32 \textcolor{blue}{(0.76)}                      & 16.8 \textcolor{blue}{(3.92)}      & \textbf{\underline{1.54}}         & 2:55    \\
\bottomrule[1pt]
\end{tabular}
}
\end{table}

\begin{table}[htbp]
\caption{
Performance metrics of different MU methods in image classification with 50\% randomly forgotten data on CIFAR-10 using ResNet-18.
}
\label{random_forgetting_50_Cifar10}
\centering
\resizebox{0.8\textwidth}{!}{
\begin{tabular}{c|cccccc}
\toprule[1pt]
\textbf{Method} & \textbf{$UA\uparrow$} & \textbf{$RA\uparrow$} & \textbf{$TA\uparrow$} & \textbf{$MIA\uparrow\downarrow$} & \textbf{$Avg.\;Gap\downarrow$} & \textbf{$RTE\downarrow$} \\
\midrule
\textbf{Retrain}                                   & 8.11    & 100.00                               & 91.62                                & 19.47                               & 0.00                                & 23:05   \\
\midrule
\textbf{FT \cite{warnecke2021machineunlearning}}   & 0.44 \textcolor{blue}{(7.67)}   & \textbf{\underline{99.96}} \textcolor{blue}{(0.04)}   & 94.23 \textcolor{blue}{(2.61)}     & 2.15 \textcolor{blue}{(17.32)}     & 6.91                                & 1:19    \\
\textbf{RL \cite{golatkar2020eternal}}             & 4.80 \textcolor{blue}{(3.31)}   & 99.55 \textcolor{blue}{(0.45)}                      & \textbf{\underline{91.31}} \textcolor{blue}{(0.31)}   & 41.95 \textcolor{blue}{(22.48)}    & 6.64                                & 2:39    \\
\textbf{GA \cite{thudi2022auditableunlearning}}     & 0.40 \textcolor{blue}{(7.71)}   & 99.61 \textcolor{blue}{(0.39)}                      & 94.34 \textcolor{blue}{(2.72)}     & 1.22 \textcolor{blue}{(18.25)}     & 7.27                                & 0:40    \\
\textbf{IU \cite{koh2017influencefunctions}}        & 3.97 \textcolor{blue}{(4.14)}   & 96.21 \textcolor{blue}{(3.79)}                      & 90.00 \textcolor{blue}{(1.62)}     & 7.29 \textcolor{blue}{(12.18)}     & 5.43                                & 3:15    \\
\textbf{BE \cite{chen2023boundaryunlearning}}       & 3.08 \textcolor{blue}{(5.03)}   & 96.84 \textcolor{blue}{(3.16)}                      & 90.41 \textcolor{blue}{(1.21)}     & 24.87 \textcolor{blue}{(5.4)}      & 3.70                                & 1:19    \\
\textbf{BS \cite{chen2023boundaryunlearning}}       & 9.76 \textcolor{blue}{(1.65)}   & 90.19 \textcolor{blue}{(8.81)}                      & 83.71 \textcolor{blue}{(7.91)}     & 32.15 \textcolor{blue}{(12.68)}    & 7.76                                & 2:07    \\
\textbf{$\ell_1$-Sparse \cite{koh2017influencefunctions}} & 1.44 \textcolor{blue}{(6.67)}   & 99.52 \textcolor{blue}{(0.48)}                      & 93.13 \textcolor{blue}{(1.51)}     & 4.76 \textcolor{blue}{(14.71)}     & 5.84                                & 1:19    \\
\textbf{SalUn \cite{fan2023salun}}                  & 7.52 \textcolor{blue}{(0.59)}   & 95.17 \textcolor{blue}{(4.83)}                      & 89.44 \textcolor{blue}{(2.18)}     & 15.96 \textcolor{blue}{(3.51)}     & 2.79                                & 2:41    \\
\midrule
\textbf{CL}                                        & 7.73 \textcolor{blue}{(0.38)}   & 96.22 \textcolor{blue}{(3.78)}                      & 90.07 \textcolor{blue}{(1.55)}     & 17.79 \textcolor{blue}{(1.68)}     & 1.85                                & 2:43    \\
\textbf{WS-CL}                                     & \textbf{\underline{8.20}} \textcolor{blue}{(0.09)} & 95.16 \textcolor{blue}{(4.84)}                      & 89.07 \textcolor{blue}{(2.55)}     & \textbf{\underline{19.72}} \textcolor{blue}{(0.25)} & 1.93                     & 2:44    \\
\textbf{WSS-KCL}                                   & 7.49 \textcolor{blue}{(0.62)}   & 96.28 \textcolor{blue}{(3.72)}                      & 90.31 \textcolor{blue}{(1.31)}     & 18.94 \textcolor{blue}{(0.53)}     & \textbf{\underline{1.54}}         & 2:44    \\
\bottomrule[1pt]
\end{tabular}
}
\end{table}

\begin{table}[htbp]
\caption{
Performance metrics of different MU methods in image classification with 10\% randomly forgotten data on CIFAR-100 using ResNet-18.
}
\label{random_forgetting_10_Cifar100}
\centering
\resizebox{0.8\textwidth}{!}{
\begin{tabular}{c|cccccc}
\toprule[1pt]
\textbf{Method} & \textbf{$UA\uparrow$} & \textbf{$RA\uparrow$} & \textbf{$TA\uparrow$} & \textbf{$MIA\uparrow\downarrow$} & \textbf{$Avg.\;Gap\downarrow$} & \textbf{$RTE\downarrow$} \\
\midrule
\textbf{Retrain}                                    & 23.51                            & 99.98                               & 74.43                              & 49.76                              & 0.00                               & 41:22   \\
\midrule
\textbf{FT \cite{warnecke2021machineunlearning}}    & 2.42 \textcolor{blue}{(21.09)}   & 99.95 \textcolor{blue}{(0.03)}     & 75.55 \textcolor{blue}{(1.12)}    & 11.04 \textcolor{blue}{(38.72)}   & 15.24                              & 2:16    \\
\textbf{RL \cite{golatkar2020eternal}}             & 55.03 \textcolor{blue}{(31.52)}  & 99.81 \textcolor{blue}{(0.17)}     & 70.03 \textcolor{blue}{(4.40)}    & 98.97 \textcolor{blue}{(49.21)}   & 21.33                              & 2:07    \\
\textbf{GA \cite{thudi2022auditableunlearning}}     & 3.13 \textcolor{blue}{(20.38)}   & 97.33 \textcolor{blue}{(2.65)}     & 75.31 \textcolor{blue}{(0.88)}    & 7.24  \textcolor{blue}{(42.52)}   & 16.61                              & 0:08    \\
\textbf{IU \cite{koh2017influencefunctions}}        & 3.18 \textcolor{blue}{(20.33)}   & 97.15 \textcolor{blue}{(2.83)}     & 73.49 \textcolor{blue}{(0.94)}    & 9.62  \textcolor{blue}{(40.14)}   & 16.06                              & 3:49    \\
\textbf{BE \cite{chen2023boundaryunlearning}}       & 2.31 \textcolor{blue}{(21.20)}   & 97.27 \textcolor{blue}{(2.71)}     & \textbf{\underline{73.93}} \textcolor{blue}{(0.50)} & 9.62  \textcolor{blue}{(40.14)}   & 17.18                              & 0:15    \\
\textbf{BS \cite{chen2023boundaryunlearning}}       & 2.27 \textcolor{blue}{(21.24)}   & 97.41 \textcolor{blue}{(2.57)}     & 75.26 \textcolor{blue}{(0.86)}    & 5.82  \textcolor{blue}{(43.94)}   & 17.15                              & 0:26    \\
\textbf{$\ell_1$-Sparse \cite{koh2017influencefunctions}} & 10.64 \textcolor{blue}{(12.87)}  & 96.62 \textcolor{blue}{(3.36)}     & 70.99 \textcolor{blue}{(3.44)}    & 22.58 \textcolor{blue}{(27.18)}   & 11.71                              & 2:17    \\
\textbf{SalUn \cite{fan2023salun}}                  & 40.29 \textcolor{blue}{(16.78)}  & 99.71 \textcolor{blue}{(0.28)}     & 69.35 \textcolor{blue}{(5.08)}    & 84.49 \textcolor{blue}{(34.73)}   & 14.22                              & 2:07    \\
\midrule
\textbf{CL}                                        & 25.44 \textcolor{blue}{(1.93)}   & 99.46 \textcolor{blue}{(0.52)}     & 69.24 \textcolor{blue}{(5.19)}    & \textbf{\underline{46.78}} \textcolor{blue}{(2.98)} & 2.66                              & 2:17    \\
\textbf{WS-CL}                                     & \textbf{\underline{23.36}} \textcolor{blue}{(0.15)} & 99.82 \textcolor{blue}{(0.16)} & 69.86 \textcolor{blue}{(4.57)}    & 43.40 \textcolor{blue}{(6.36)}   & 2.81                              & 2:19    \\
\textbf{WSS-CL}                                    & 25.05 \textcolor{blue}{(1.13)}   & \textbf{\underline{99.95}} \textcolor{blue}{(0.03)} & 69.65 \textcolor{blue}{(4.78)}    & 45.62 \textcolor{blue}{(4.14)}   & \textbf{\underline{2.53}}         & 2:19    \\
\bottomrule[1pt]
\end{tabular}
}
\end{table}

\begin{table}[htbp]
\caption{
Performance metrics of different MU methods in image classification with 50\% randomly forgotten data on CIFAR-100 using ResNet-18.
}
\label{random_forgetting_50_Cifar100}
\centering
\resizebox{0.8\textwidth}{!}{
\begin{tabular}{c|cccccc}
\toprule[1pt]
\textbf{Method} & \textbf{$UA\uparrow$} & \textbf{$RA\uparrow$} & \textbf{$TA\uparrow$} & \textbf{$MIA\uparrow\downarrow$} & \textbf{$Avg.\;Gap\downarrow$} & \textbf{$RTE\downarrow$} \\
\midrule
\textbf{Retrain}                                    & 33.24                            & 99.98                               & 67.18                              & 61.43                              & 0.00                               & 41:22   \\
\midrule
\textbf{FT \cite{warnecke2021machineunlearning}}    & 2.71 \textcolor{blue}{(30.53)}   & 99.96 \textcolor{blue}{(0.02)}     & 75.11 \textcolor{blue}{(7.93)}    & 10.71 \textcolor{blue}{(50.72)}   & 22.30                              & 1:15    \\
\textbf{RL \cite{golatkar2020eternal}}             & 50.52 \textcolor{blue}{(17.28)}  & 99.47 \textcolor{blue}{(0.51)}     & 56.75 \textcolor{blue}{(10.43)}   & 95.91 \textcolor{blue}{(34.48)}   & 15.68                              & 2:08    \\
\textbf{GA \cite{thudi2022auditableunlearning}}     & 2.61 \textcolor{blue}{(31.08)}   & 97.49 \textcolor{blue}{(2.49)}     & 75.27 \textcolor{blue}{(8.09)}    & 5.92  \textcolor{blue}{(55.51)}   & 24.29                              & 0:40    \\
\textbf{IU \cite{koh2017influencefunctions}}        & 12.64 \textcolor{blue}{(20.60)}  & 87.96 \textcolor{blue}{(12.02)}    & 62.76 \textcolor{blue}{(4.42)}    & 10.32 \textcolor{blue}{(51.11)}   & 22.04                              & 3:48    \\
\textbf{BE \cite{chen2023boundaryunlearning}}       & 2.76 \textcolor{blue}{(30.48)}   & 97.39 \textcolor{blue}{(2.59)}     & 74.05 \textcolor{blue}{(6.87)}    & 7.44  \textcolor{blue}{(53.99)}   & 23.48                              & 1:16    \\
\textbf{BS \cite{chen2023boundaryunlearning}}       & 2.99 \textcolor{blue}{(30.25)}   & 97.24 \textcolor{blue}{(2.74)}     & 73.38 \textcolor{blue}{(6.20)}    & 7.63  \textcolor{blue}{(53.80)}   & 16.50                              & 2:05    \\
\textbf{$\ell_1$-Sparse \cite{koh2017influencefunctions}} & 39.86 \textcolor{blue}{(6.62)}   & 78.17 \textcolor{blue}{(21.81)}    & 57.66 \textcolor{blue}{(9.52)}    & 40.21 \textcolor{blue}{(21.22)}   & 14.78                              & 1:16    \\
\textbf{SalUn \cite{fan2023salun}}                  & 47.07 \textcolor{blue}{(13.83)}  & 99.79 \textcolor{blue}{(0.19)}     & 56.64 \textcolor{blue}{(17.79)}   & 85.17 \textcolor{blue}{(23.74)}   & 13.89                              & 2:07    \\
\midrule
\textbf{CL}                                        & \textbf{\underline{32.78}} \textcolor{blue}{(0.46)} & \textbf{\underline{99.98}} \textcolor{blue}{(0.00)} & 64.49 \textcolor{blue}{(2.49)} & 45.83 \textcolor{blue}{(15.60)} & 4.64                              & 2:17    \\
\textbf{WS-CL}                                     & 32.63 \textcolor{blue}{(0.61)}  & 99.67 \textcolor{blue}{(0.31)}     & 64.52 \textcolor{blue}{(2.66)}   & 44.33 \textcolor{blue}{(17.10)}   & 5.17                              & 2:19    \\
\textbf{WSS-CL}                                    & 32.25 \textcolor{blue}{(0.99)}  & \textbf{\underline{99.97}} \textcolor{blue}{(0.01)} & \textbf{\underline{66.45}} \textcolor{blue}{(0.73)} & \textbf{\underline{49.38}} \textcolor{blue}{(12.05)} & \textbf{\underline{3.44}} & 2:19    \\
\bottomrule[1pt]
\end{tabular}
}
\end{table}

\begin{table}[ht]
\centering
\caption{Comparison of the original model, retrained model, WSS-CL, and SalUn methods in the per-class forgetting task; each model is evaluated by four metrics across the 10 classes of the CIFAR-10 dataset}
\label{two_model_five_methods}
\resizebox{0.9\textwidth}{!}{
\begin{tabular}{l|ccc|ccc|ccc|ccc}
\toprule
& \multicolumn{3}{c|}{\textbf{Original}} & \multicolumn{3}{c|}{\textbf{Retrain}} & \multicolumn{3}{c|}{\textbf{WSS-CL}} & \multicolumn{3}{c}{\textbf{SalUn} \cite{fan2023salun}} \\
\midrule

\textbf{Metric} & \textbf{$UA \uparrow$} & \textbf{$RA \uparrow$} & \textbf{$TA \uparrow$} 
                & \textbf{$UA \uparrow$} & \textbf{$RA \uparrow$} & \textbf{$TA \uparrow$} 
                & \textbf{$UA \uparrow$} & \textbf{$RA \uparrow$} & \textbf{$TA \uparrow$} 
               & \textbf{$UA \uparrow$} & \textbf{$RA \uparrow$} & \textbf{$TA \uparrow$} \\
\midrule
\textbf{airplane} & 0.444 & 99.86 & 94.60 & 100.0 & 99.86 & 92.10 &  \textbf{\underline{100.0}} & 99.37 & 93.63 & 99.87 & 99.56 & 93.83 \\
\textbf{automobile} & 0.556 & 99.98 & 93.7 & 100.0 & 99.99 & 91.84 &  100.0 & 98.88 & 92.4 & 100.0 & 99.23 & 92.78 \\
\textbf{bird}  & 0.889 & 99.99 & 94.60   & 100.0 & 100.0 & 93.07  &  \textbf{\underline{100.0}} & 99.13 & 93.77 & 95.98 & 99.76 & 94.7 \\
\textbf{cat} & 0.578 & 100.0 & 94.63   & 100.0 & 100.0 & 93.88  &  100.0 & 99.80 & 95.39 & 100.0 & 99.80 & 95.57 \\
\textbf{deer}  & 0.622 & 100.0 & 94.51 & 100.0 & 99.99 & 92.72 &  100.0 & 99.37 & 93.8 & 100.0 & 99.60 & 94.07 \\
\textbf{dog}   & 0.956 & 99.96 & 94.91  & 100.0 & 100.0 & 93.23 &  \textbf{\underline{100.0}} & 99.80 & 95.07 & 99.93 & 99.84 & 95.16 \\
\textbf{frog}  & 0.478 & 100.0 & 94.91  & 100.0 & 100.0 & 92.11  &  100.0 & 98.42 & 92.18 & 100.0 & 99.54 & 93.56  \\
\textbf{horse}   & 0.244 & 100.0 & 94.48  & 100.0 & 100.0 & 92.02 &  100.0 & 98.87 & 92.61 & 100.0 & 99.64 & 93.63 \\
\textbf{ship}   & 0.244 & 99.996 & 94.40 & 100.0 & 100.0 & 91.79  &  \textbf{\underline{100.0}} & 99.33 & 92.97 & 97.62 & 99.49 & 93.19 \\
\textbf{truck}   & 0.578 & 100.0 & 94.37  & 100.0 & 100.0 & 92.24 &  100.0 & 98.35 & 91.8 & 100.0 & 99.52 & 93.6 \\

\bottomrule
\end{tabular}
}
\end{table}

Table~\ref{random_forgetting_10_Cifar10} illustrates performance of several MU algorithms when 10\% of training examples are randomly forgotten. Metrics considered include UA, RA, TA, MIA, Avg. Gap, and RTE. 

Method WSS-CL attains a minimum Avg. Gap of 1 minute 32 seconds, signifying its best approximation for the ideal unlearning algorithm, Retain. The WS-CL alternative comes in at a narrow margin with an Avg. Gap of 1 minute 33 seconds, proving its efficiency, albeit slightly less. For comparison, baseline algorithms such as FT and GA have larger Avg. Gaps of 3 minutes 47 seconds and 4 minutes 7 seconds, respectively.

Method WSS-CL holds high unlearning accuracy (UA) at 4.51, retrain accuracy (RA) at 99.39, and test accuracy (TA) at 93.32, balancing unlearning efficacy with model performance. Although SalUn yields a relatively high UA at 4.18, its reduced RA (99.13) and TA (93.17) reveal a model fidelity penalty. In the membership inference attack (MIA), SalUn attained a value of 14.39, which is even larger than baseline values such as FT (2.70), but in comparison with the WSS-CL model (16.8), WSS-CL proves to be better at estimating through MIA. Regarding runtime efficiency (RTE), SalUn has competitive computational efficiency with an RTE value of 2 minutes 40 seconds, while our model provides 2 minutes 55 seconds. This is appreciably less compared to Retain's RTE value of 43 minutes 17 seconds, and our work proves practically beneficial in such cases.

When escalating the unlearning requirement to 50\%, Table~\ref{random_forgetting_50_Cifar10} shows the performance of various MU methods on CIFAR-10 with 50\% randomly forgotten data. For the three models, CL achieves an Avg. Gap of 1.85 with RA of 96.22 and TA of 90.07, WS-CL yields an Avg. Gap of 1.93 with RA of 95.16 and TA of 89.07, and notably, WSS-CL attains the lowest Avg. Gap of 1.54 along with the highest RA (96.28) and TA (90.31), indicating its superior balance between unlearning efficacy and model fidelity; all three methods exhibit comparable runtime efficiency (approximately 2 minutes 44 seconds).

We perform dropout of 10\% on the CIFAR-100 dataset in Table~\ref{random_forgetting_10_Cifar100} and 50\% on the CIFAR-100 dataset in Table~\ref{random_forgetting_50_Cifar100}, which shows that WSS-CL performs well, with an average gap of 2.53 at a 10\% dropout rate and 3.44 at a 50\% dropout rate.

\subsection{Discussion and Insights}

Our proposed algorithm, \textit{Weight Saliency Soft-Guided Contrastive Learning (WSS-CL)}, has a number of key improvements over its gradient-based weight saliency counterpart in \cite{fan2023salun} in terms of unlearning in machines. These improvements occur through our new two-step mechanism with a bias towards effective forgetting and fine-tuning, and with less use of external randomness.

\textbf{Effective forgetting through Kullback-Leibler in logit space}. In forgetting, our algorithm optimizes for the Kullback-Leibler divergence between model output logits and uniform pseudo-labels for unlearned (forgotten) samples. It formulates unlearning in logit space explicitly, such that forgetting is explicitly tied to model output behavior, with a principled and deterministic forgetting mechanism.

\textbf{Adversarial fine-tuning with self-supervised contrastive learning}. Our algorithm utilizes a new loop forgetting mechanism in its fine-tuning through the use of self-supervised contrastive learning. By maximizing the distance between representations for unlearned and retained samples, our algorithm ensures the effective erasure of unlearned data’s contribution. In contrast, with salUn, the work involves reassigning labels at a high level and not with feature-space separation explicitly.

\textbf{Adjust the saliency masking method with saliency masking soft}. Soft saliency masking improves upon conventional saliency masking by providing a continuous measure of importance for each parameter. This approach preserves more gradient information, avoids harsh thresholds, and provides a smoother, more robust mechanism for targeting model parameters during unlearning. The result is a method that not only effectively reduces the effects of forgetting but also maintains better performance in retention.

\section{Conclusion}

In this work, we circumvent the weaknesses of present unlearning techniques with the new principle of weight saliency, using effective forgetting through KL divergence in logit space and adversarial fine-tuning with contrastive learning. A new unlearning scheme arises. Our algorithm effectively overcomes all of the weaknesses of present techniques and is best suited for image classification environments in general. As a compelling use case, WSS-CL can effectively prevent stable diffusion networks from generating offensive images, even when prompted with offensive prompts to generate an image. With the significant increase in efficiency and performance, the value of weight-saliency soft unlearning is emphasized, and future work will, in our view, generalize and build in a variety of directions with refinements. Our future work may apply the WSS-CL architecture to problems such as preventing the generation of harmful images, such as faces and emotions.
%
% ---- Bibliography ----
%
% BibTeX users should specify bibliography style 'splncs04'.
% References will then be sorted and formatted in the correct style.
%
% \bibliographystyle{splncs04}
\bibliographystyle{splncs04}         
\bibliography{iclr2025_conference}   

\end{document}